\pdfoutput=1
\PassOptionsToPackage{table}{xcolor}
\documentclass[11pt]{article}

\usepackage[preprint]{acl}

\usepackage{times}
\usepackage{latexsym}

\usepackage[T1]{fontenc}

\usepackage[utf8]{inputenc}

\usepackage{microtype}

\usepackage{inconsolata}

\usepackage{graphicx}

\usepackage{booktabs}

\usepackage{CJKutf8}
\usepackage{enumitem}

\usepackage[most]{tcolorbox}
\definecolor{skyblue}{RGB}{95, 157, 241}

\usepackage{multirow}

\title{From Monolingual to Bilingual: Investigating Language Conditioning in Large Language Models for Psycholinguistic Tasks}

\author{Shuzhou Yuan,~Zhan Qu,~Mario Tawfelis,
~and Michael Färber \\
ScaDS.AI and TU Dresden \\
\texttt{\{shuzhou.yuan, zhan.qu\}@tu-dresden.de}}

\begin{document}
\maketitle
\begin{abstract}
Large Language Models (LLMs) exhibit strong linguistic capabilities, but little is known about how they encode psycholinguistic knowledge across languages. We investigate whether and how LLMs exhibit human-like psycholinguistic responses under different linguistic identities using two tasks: sound symbolism and word valence. We evaluate two models, \texttt{Llama-3.3-70B-Instruct} and \texttt{Qwen2.5-72B-Instruct}, under monolingual and bilingual prompting in English, Dutch, and Chinese. Behaviorally, both models adjust their outputs based on prompted language identity, with Qwen showing greater sensitivity and sharper distinctions between Dutch and Chinese. Probing analysis reveals that psycholinguistic signals become more decodable in deeper layers, with Chinese prompts yielding stronger and more stable valence representations than Dutch. Our results demonstrate that language identity conditions both output behavior and internal representations in LLMs, providing new insights into their application as models of cross-linguistic cognition.

\end{abstract}

\section{Introduction}

As Large Language Models (LLMs) continue to advance and as generative AI becomes increasingly integrated into everyday life \citep{achiam2023gpt,liu2024deepseek}, researchers have turned to psycholinguistic paradigms to investigate their alignment with human cognition and behavior \citep{hagendorff2023machine,liu2025mind,yuan2025hateful}. Studies have found that LLMs often exhibit human-like intuitions in language processing, though with systematic divergences \citep{lee-etal-2024-psycholinguistic,amouyal-etal-2024-large,trott2024can}. However, most prior experiments were conducted in English, leaving open important questions about how LLMs behave across languages, and how language identity, especially in bilingual or multilingual contexts, shapes their psycholinguistic behavior. 

\begin{figure}[ht]
    \centering
    \includegraphics[width=0.9\linewidth]{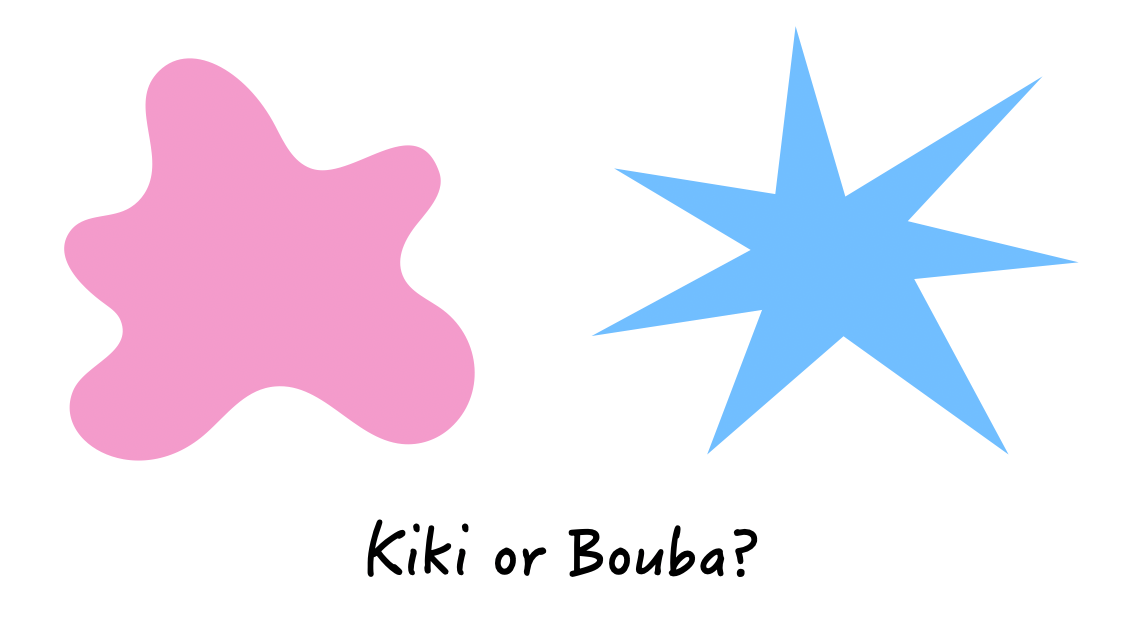}
    \caption{The Bouba-Kiki effect illustrates that people tend to associate round shapes with the nonsense word “Bouba” and spiky shapes with “Kiki,” reflecting a non-arbitrary link between sound and visual form.}
    \label{figure:intro_figure}
\end{figure}


Research in both cognitive science and machine learning has shown that the relationship between sounds and meanings is not entirely arbitrary \citep{alper2023kiki,fort2022resolving}. A well-known example is the Bouba–Kiki effect, illustrated in Figure~\ref{figure:intro_figure}, where speakers consistently associate rounded shapes with the word "bouba" and spiky shapes with "kiki". Beyond this, sound symbolism, where specific phonetic patterns consistently evoke particular semantic or affective impressions, has been observed across languages, particularly among bilingual individuals \citep{louwerse2017estimating}. For instance, native Dutch speakers tend to perceive unfamiliar Chinese words with certain phonological patterns (nasal-initial words) as negative, but native Chinese speakers tend to perceive unfamiliar Dutch words with the same phonological patterns as positive. These cross-linguistic intuitions provide a compelling foundation for investigating whether LLMs simulate similar effects under different linguistic identities.

\begin{figure*}[ht]
    \centering
    \includegraphics[width=0.85\linewidth]{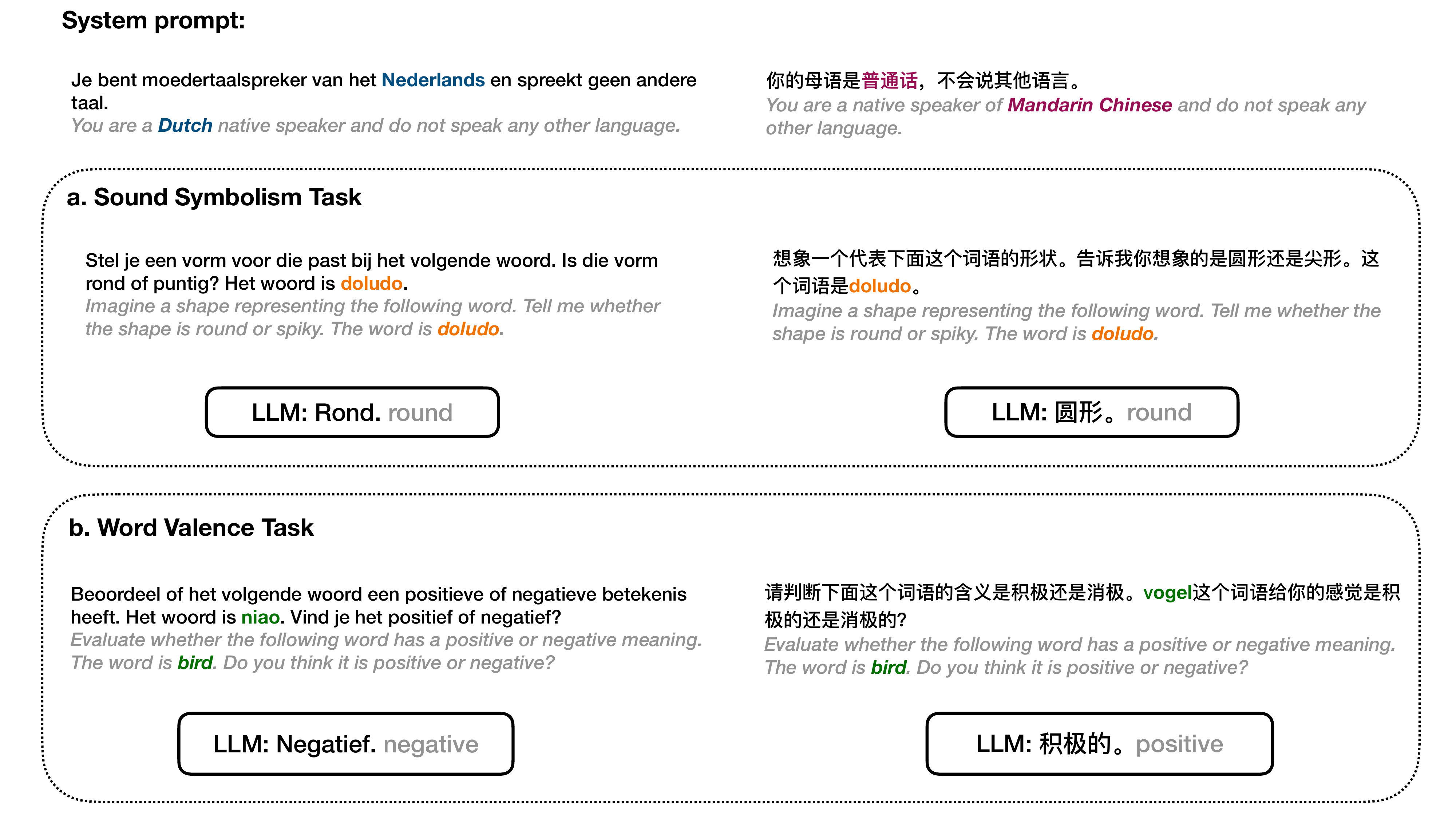}
    \caption{Overview of two psycholinguistic tasks. 
    (a) \textbf{Sound Symbolism Task:} The LLM is prompted to judge whether a pseudoword (e.g., \textit{doludo}) evokes a round or spiky shape. 
    (b) \textbf{Word Valence Task:} The LLM is asked to determine whether a real word (e.g., \textit{vogel} / \textit{niao}) has a positive or negative connotation. 
    Prompts are presented in Dutch and Mandarin for monolingual conditions. For bilingual conditions, the system prompt is in English while the user message remains in Dutch or Chinese.}
    \label{figure:overview}
\end{figure*}

Inspired by \citet{louwerse2017estimating}, we evaluate LLMs on two psycholinguistic tasks: (1) a \textbf{sound symbolism} task, in which models classify pseudowords as evoking a round or spiky shape, and (2) a \textbf{word valence} task, in which we test whether real words are perceived as emotionally positive or negative.\footnote{While word valence is a type of sound symbolism, we use this term to distinguish it from the other task.} As shown in Figure~\ref{figure:overview}, we probe model behavior in four settings: Dutch and Chinese monolingual prompts, and Dutch-English and Chinese-English bilingual prompts. The sound symbolism task is based on pseudowords annotated for shape \citep{alper2023kiki}, while the word valence task uses real English words labeled by affective norms \citep{bradley1999affective}, translated into Dutch and Chinese. Translation quality was manually verified and cross-checked against the data from \citet{louwerse2017estimating}.

We compare two models of comparable scale: \texttt{Llama-3.3-70B-Instruct} \citep{grattafiori2024llama}, which is trained primarily on English data without explicit coverage of Dutch or Chinese, and \texttt{Qwen2.5-72B-Instruct} \citep{qwen2025qwen25technicalreport}, a multilingual model trained on both Dutch and Chinese. We find that both models adapt their outputs based on the linguistic identity embedded in the prompt. While \texttt{Llama} exhibits relatively consistent behavior favoring monolingual prompts, \texttt{Qwen} displays stronger sensitivity to the prompt language and more pronounced behavioral divergence between Dutch and Chinese contexts, especially under bilingual prompting. These results suggest that multilingual models may simulate language-specific perceptual patterns more flexibly, though not always in alignment with English-based ground truth.


To study whether these behavioral effects correspond to internal model representations, we conduct layer-wise probing on \texttt{Llama-3.3-70B} using a simple MLP classifier trained on hidden states. Probing results show that both sound-symbolic and valence-related signals become increasingly decodable in deeper layers. Notably, word valence representations are stronger and more stable under Chinese prompts than Dutch, in both monolingual and bilingual settings. These findings support the hypothesis that LLMs internalize language-dependent cognitive patterns and that prompt conditioning can modulate not only output behavior but also internal encoding.

This work makes the following contributions:
\begin{itemize}[noitemsep, topsep=0pt]
    \item We design two psycholinguistic tasks, sound symbolism and word valence, and are the first to evaluate LLMs under both monolingual and bilingual language conditions in this context.
    \item We conduct systematic experiments across three languages: English, Dutch, and Chinese, using two LLMs with contrasting training profiles, examining how language identity shapes model behavior.
    \item We perform layer-wise probing to examine how psycholinguistic information is encoded in internal representations across linguistic conditions.
\end{itemize}

Our findings offer new insights into cross-linguistic cognitive alignment in LLMs and highlight the importance of prompt conditioning in multilingual psycholinguistic modeling.

\section{Related Work}

\paragraph{LLMs in Psycholinguistics and Sound Symbolism.}

Sound symbolism refers to systematic, non-arbitrary mappings between word forms and meanings, such as the tendency to associate certain sounds with shapes, sizes, or affective valence \citep{dingemanse2016sound,ADELMAN2018122,moorthy-etal-2018-nike,BOTTINI201962,louwerse2008embodied,abramova-fernandez-2016-questioning}. Cross-linguistic studies have shown that these associations are robust \citep{blasi2016sound}, though they are shaped by the phonological background of the speaker \citep{louwerse2017estimating}. For instance, \citet{louwerse2017estimating} demonstrate that bilingual speakers rely on the phonological cues of their native language when evaluating unfamiliar words in another language, suggesting an interaction between sound symbolism and linguistic identity.

Recent work has begun to investigate LLMs through the lens of cognitive modeling, probing their ability to replicate psycholinguistic phenomena \citep{giulianelli-etal-2024-proper,duan-etal-2025-unveiling,cong-2022-psycholinguistic,amouyal-etal-2024-large,conde2025psycholinguistic}. \citet{loakman-etal-2024-ears} show that LLMs exhibit varying levels of agreement with human annotations in sound symbolism tasks, while \citet{verhoef-etal-2024-kiki} find that multimodal models encode the bouba–kiki effect in a manner consistent with human perception. Similarly, \citet{alper2023kiki} probe these models and find strong evidence that they encode sound-symbolic patterns aligning with those observed in human studies. Other studies leverage LLMs to augment psycholinguistic datasets and evaluate sensitivity to linguistic cues in language understanding tasks \citep{trott2024can,lee-etal-2024-psycholinguistic}.

Despite these advances, little attention has been paid to how language conditioning, particularly monolingual versus bilingual prompting, modulates LLM behavior in sound symbolism contexts. Our work addresses this gap by systematically investigating how language identity affects both model output and internal representations in psycholinguistic tasks.

\paragraph{Probing and Representation Analysis.}
Probing methods are widely used to assess whether LLMs encode specific linguistic or conceptual properties in their internal representations \citep{weissweiler-etal-2022-better,vulic-etal-2020-probing,pimentel-etal-2020-information,ousidhoum-etal-2021-probing}. \citet{arora-etal-2023-probing} examine how cultural values are embedded in LLMs, while \citet{koto-etal-2021-discourse} propose discourse-level probes to capture long-range textual relationships. \citet{wang-etal-2024-probing-emergence} explore cross-lingual neuron overlap and its implications for zero-shot transfer, and \citet{roy-etal-2023-probing} investigate internal representations related to hate speech detection.

Probing has also been applied to psycholinguistic data. \citet{bazhukov-etal-2024-models} and \citet{shivagunde-etal-2023-larger} evaluate the degree to which LLMs align with human cognitive benchmarks. However, existing work typically assumes a fixed language context. Our approach extends this line of research by examining how psycholinguistic representations, specifically sound symbolism and word valence, shift with language identity, offering new insights into multilingual cognition in LLMs.

\section{Psycholinguistics Tasks}

We design two psycholinguistic tasks, \textit{sound symbolism} and \textit{word valence}, to evaluate the behavior of LLMs across linguistic contexts. The dataset statistics are summarized in Table~\ref{tab:dataset_statistics}. Following prior work on cross-linguistic psycholinguistic evaluation \citep{louwerse2017estimating}, we focus on Dutch and Chinese as our primary languages of interest.

\vspace{-0.1cm}
\begin{table}[ht]
\centering
\footnotesize
\begin{tabular}{lll}
\toprule
\textbf{Task} & \textbf{Number} & \textbf{Label}  \\
\midrule
Sound symbolism  & 648  & Round, Spiky\\
Word valence   &   1034 & Positive, Negative\\
\bottomrule
\end{tabular}
\caption{Statistics of the two psycholinguistics tasks.}
\label{tab:dataset_statistics}
\end{table}
\vspace{-0.6cm}

\subsection{Task 1: Sound Symbolism}


We base our first task on the dataset introduced by \citet{alper2023kiki}, which contains 648 pseudowords annotated with corresponding shape labels: \textit{round} or \textit{spiky}. This dataset captures the well-documented Bouba–Kiki effect, where particular phonetic features are consistently linked to specific visual forms. 

In this task, we prompt the LLM with a pseudoword and ask whether it evokes a round or spiky shape. This allows us to assess whether the model exhibits sensitivity to sound-shape associations in a manner consistent with human perception.



\subsection{Task 2: Word Valence}



To evaluate semantic-affective judgments, we use a set of real English words from the Affective Norms for English Words (ANEW) dataset \citep{bradley1999affective}. Each word is labeled as either \textit{positive} or \textit{negative} based on its mean human-rated valence score. Following \citet{louwerse2017estimating}, we translate these words into Dutch and Chinese using the Google Translate API. For Chinese words, we additionally provide their romanized forms (Pinyin) to enable cross-linguistic comparison based on phonological features.


This multilingual extension enables us to assess how LLMs interpret emotional valence across languages. The task simulates a scenario in which a speaker encounters a word from an unfamiliar language and infers its affective meaning based on phonological cues. As shown in Figure~\ref{figure:overview}, we test whether the LLM's responses vary depending on its assigned linguistic identity, for instance, presenting a Dutch word to a `Chinese' LLM persona, or a romanized Chinese word to a `Dutch' one.

\section{Experiments}

\subsection{Models}


We evaluate two state-of-the-art open-source LLMs with approximately 70 billion parameters: \texttt{Llama-3.3-70B-Instruct} \citep{grattafiori2024llama} and \texttt{Qwen2.5-72B-Instruct} \citep{qwen2025qwen25technicalreport}. While \texttt{Llama-3.3} was trained primarily on English and lacks explicit exposure to Dutch and Chinese, \texttt{Qwen2.5} includes both languages in its training corpus, making it explicitly multilingual. Prior studies suggest that even models with English-centric training can exhibit emergent multilingual abilities without direct language supervision \citep{nie2024decomposed}. This model selection enables a systematic comparison between an English-dominant model and a multilingual one across psycholinguistic tasks under monolingual and bilingual conditions. All generations are produced using a temperature of 0 to ensure deterministic outputs.

\subsection{Prompt Formulation}

To simulate different linguistic identities, we embed persona information in the system prompt, following prior work on prompt-based conditioning \citep{yuan2025hateful}.

For the \textbf{monolingual} setting, we use the following system prompt:

\begin{tcolorbox}[colback=skyblue!20, colframe=skyblue!80!black, width=0.50\textwidth, rounded corners, boxsep=2pt, left=2pt, right=2pt, top=2pt, bottom=2pt]
\textit{System}: You are a native speaker of \texttt{[language]} and do not speak any other language.
\end{tcolorbox}

Here, \texttt{[language]} denotes either `Mandarin Chinese' or `Standard Dutch', with both the system and user prompts consistently written in that language to simulate a monolingual environment.

For the \textbf{bilingual} setting, where English is always paired with Dutch or Chinese, the system prompt is:

\begin{tcolorbox}[colback=skyblue!20, colframe=skyblue!80!black, width=0.50\textwidth, rounded corners, boxsep=2pt, left=2pt, right=2pt, top=2pt, bottom=2pt]
\textit{System}: You are a bilingual speaker of English and \texttt{[language]}, and do not speak any other language.
\end{tcolorbox}

In this case, the prompt is written in English, while user input remains in Dutch or Chinese.








\subsection{Results and Analysis}

\begin{table}[ht]
\centering
\scalebox{0.8}{
\begin{tabular}{llllll}
\toprule
\textbf{Model}   & \multicolumn{2}{c}{\textbf{Language profile}} & \textbf{Task 1}  & \textbf{Task 2} & \textbf{Avg.} \\
\midrule
\multirow{5}{*}{\texttt{Llama}}     & \multirow{3}{*}{Monolingual} &  \cellcolor{gray!20}English     &  \cellcolor{gray!20}70.52 &  \cellcolor{gray!20}99.17 & \cellcolor{gray!20}84.85 \\
                           &    &   Dutch      &  64.81 &  56.17 & 60.49 \\
                           &    &   Chinese    &  60.65 &  86.67 & 73.66 \\
\cline{2-6}
                           & \multirow{2}{*}{Bilingual} & Dutch   &  51.08 & 49.67 & 50.38 \\
                           &    &  Chinese               &  50.31 & 82.83 & 66.57 \\
\midrule
\midrule
\multirow{5}{*}{\texttt{Qwen}}     & \multirow{3}{*}{Monolingual} &  \cellcolor{gray!20}English     & \cellcolor{gray!20}57.87 & \cellcolor{gray!20}96.33 & \cellcolor{gray!20}77.10 \\
                           &    &   Dutch      & 66.20 & 21.00 & 43.60 \\
                           &    &   Chinese    & 13.43 & 35.50 & 24.47 \\
\cline{2-6}
                           & \multirow{2}{*}{Bilingual} & Dutch   & 66.51 & 67.50 & 67.01 \\
                           &    &  Chinese               & 19.29 & 25.83 & 22.56 \\
\bottomrule
\end{tabular}}
\caption{Accuracy (\%) of \texttt{Llama} and \texttt{Qwen} compared to English-based ground truth labels. English monolingual results serve as reference points, as Task 2 labels are derived from native English speakers.}
\label{tab:results_prompt}
\end{table}

Table~\ref{tab:results_prompt} presents model accuracy across tasks and language conditions. Since the ground truth labels are derived from English speakers, performance in Dutch and Chinese contexts should not be interpreted as absolute accuracy. Rather, these scores serve to illuminate how LLM behavior shifts under different linguistic identities—specifically between monolingual and bilingual prompting, and across Dutch and Chinese inputs.

In the English monolingual condition, both models perform well on the word valence task (Task 2), with \texttt{Llama} achieving 99.17\% and \texttt{Qwen} 96.33\%. This strong performance is expected, as the prompts, labels, and training data are all English-aligned. On the sound symbolism task (Task 1), however, the models diverge more substantially: \texttt{Llama} reaches 70.52\% accuracy, indicating moderate alignment with human-like phonological associations, while \texttt{Qwen} achieves only 57.87\%, suggesting weaker or less consistent sensitivity to sound-shape correspondences.

These results highlight two key observations: first, both models align well with English affective norms when operating in their strongest linguistic context; second, \texttt{Llama}, despite lacking multilingual training, exhibits more stable behavior in capturing sound symbolism under English prompts. In contrast, \texttt{Qwen}’s broader multilingual exposure may introduce variability that reduces its alignment with English-specific psycholinguistic patterns.



\subsubsection{Monolingual vs. Bilingual Prompting}

To examine how language conditioning affects model predictions, we define the discrepancy between monolingual and bilingual accuracy as:

\vspace{-0.3cm}
\begin{equation}
D_m = \text{Accuracy}_{\text{bi}} - \text{Accuracy}_{\text{mo}}
\end{equation}
\vspace{-0.3cm}

\noindent A positive \(D_m\) indicates that bilingual prompting leads to higher alignment with English-based ground truth labels, suggesting enhanced sensitivity when both English and the target language are specified. A negative \(D_m\) by contrast, suggests that monolingual prompts produce stronger alignment, implying that bilingual conditioning may weaken or blur language-specific cues.

\begin{table}[ht]
\centering
\footnotesize
\begin{tabular}{lllll}
\toprule
\textbf{Model} & \textbf{Language} & \textbf{Task 1} & \textbf{Task 2} & \textbf{Avg.} \\
\midrule
\multirow{2}{*}{\texttt{Llama}} & Dutch   & -13.73 & -6.50 & -10.12 \\
                                & Chinese & -10.34 & -3.84 & -7.09 \\
\hline
\multirow{2}{*}{\texttt{Qwen}}  & Dutch   & 0.31   & 46.50 & 23.41 \\
                                & Chinese & 5.86   & -9.67 & -1.91 \\
\bottomrule
\end{tabular}
\caption{Discrepancy $D_m$ between bilingual and monolingual prompts across tasks and languages.}
\label{tab:results_discrepancy_multilingual}
\end{table}

As shown in Table~\ref{tab:results_discrepancy_multilingual}, \texttt{Llama} consistently favors monolingual prompting, particularly for sound symbolism (Task 1), where bilingual prompting reduces accuracy by more than 10\% in both Dutch and Chinese. This suggests that adding English to the linguistic context may disrupt phonological reasoning in this model.

In contrast, \texttt{Qwen} shows more mixed behavior. For Dutch, bilingual prompting leads to a substantial improvement in Task 2 (+46.5\%) while leaving Task 1 largely unaffected (+0.31\%). For Chinese, bilingual prompting slightly improves Task 1 (+5.86) but reduces Task 2 performance (–9.67), partly due to synonymous responses being marked as incorrect under strict exact-match evaluation. For instance, \texttt{Qwen} may generate "joyful" instead of "positive", a semantically valid response, but one penalized by the evaluation metric.

These patterns highlight a tradeoff in bilingual conditioning: while it may enrich semantic associations in multilingual models like \texttt{Qwen}, it can also introduce instability or ambiguity, especially when exact lexical matches are required.

\subsubsection{Chinese vs. Dutch}

To understand how different languages influence psycholinguistic behavior in LLMs, we compute the discrepancy between Chinese and Dutch prompts under matched persona settings (monolingual or bilingual):

\vspace{-0.3cm}
\begin{equation}
D_l = \text{Accuracy}_{\text{Chinese}} - \text{Accuracy}_{\text{Dutch}}
\end{equation}
\vspace{-0.3cm}

\noindent Here, a positive \(D_l\) indicates that the model aligns more closely with the English-based ground truth when prompted in Chinese than in Dutch; a negative \(D_l\) suggests stronger alignment under Dutch prompting. These values do not represent absolute performance, but instead quantify how language identity shifts the model's alignment relative to a fixed reference.




\begin{table}[ht]
\centering
\footnotesize
\begin{tabular}{lllll}
\toprule
\textbf{Model} & \textbf{Setting} & \textbf{Task 1} & \textbf{Task 2} & \textbf{Avg.} \\
\midrule
\multirow{2}{*}{\texttt{Llama}} & Monolingual   & -4.16  & 30.50  & 13.17 \\
                                & Bilingual     & -0.77  & 33.16  & 16.20 \\
\hline
\multirow{2}{*}{\texttt{Qwen}}  & Monolingual   & -52.77 & 14.50  & -19.14 \\
                                & Bilingual     & -47.22 & -41.76 & -44.49 \\
\bottomrule
\end{tabular}
\caption{Discrepancy $D_l$ between Chinese and Dutch prompts.}
\label{tab:results_discrepancy_cn_du}
\end{table}

As shown in Table~\ref{tab:results_discrepancy_cn_du}, both models exhibit language-dependent behavior, though with markedly different magnitudes.

For \texttt{Llama}, Chinese prompts consistently improve performance in Task 2 (word valence), with large positive discrepancies of +30.50 (monolingual) and +33.16 (bilingual). In contrast, Dutch prompts slightly outperform Chinese in Task 1 (sound symbolism), resulting in modest negative values (–4.16 \& –0.77). This suggests that \texttt{Llama}’s encoding of valence is more sensitive to the Chinese phonological or semantic context, while sound-symbolic mappings are more stable or accessible through Dutch.

\texttt{Qwen} displays much stronger and more variable language effects. In Task 1, Dutch prompts lead to significantly better alignment than Chinese prompts, with large negative discrepancies of –52.77 (monolingual) and –47.22 (bilingual). However, in Task 2, Qwen’s behavior reverses in the monolingual setting, showing a moderate preference for Chinese (+14.50). This advantage disappears under bilingual prompting, where Dutch unexpectedly outperforms Chinese by a large margin (–41.76). This inversion illustrates Qwen’s heightened sensitivity to language identity, particularly under bilingual conditions, where language priors appear to compete or interfere.

Taken together, these results show that language identity plays a critical role in how LLMs simulate psycholinguistic judgments. While \texttt{Llama} maintains relatively consistent behavior with modest differences across languages, \texttt{Qwen} shows amplified and task-dependent shifts, suggesting that multilingual models are more reactive to linguistic context, even when this reactivity leads to unstable or contradictory behaviors.

\subsection{Qualitative Analysis}

\begin{CJK}{UTF8}{gbsn}
\begin{table*}[ht]
\centering
\footnotesize
\begin{tabular}{lll|l|ll|ll}
\toprule
\multicolumn{3}{c|}{\textbf{Input Word}} & \multirow{2}{*}{\textbf{Label}}  &  \multicolumn{2}{c|}{\textbf{Monolingual}} &\multicolumn{2}{c}{\textbf{Bilingual}} \\

\textbf{English} & \textbf{Dutch} & \textbf{Chinese (pinyin)} &   & \textbf{Dutch} & \textbf{Chinese} & \textbf{Dutch} & \textbf{Chinese} \\
\midrule
death       & dood         & si wang       & negative & negatief & 积极     & negatief & 积极     \\
fear        & angst        & hai pa        & negative & negatief & 消极     & positief & 消极     \\
worm        & made         & qu            & negative & negatief & 积极     & negatief & 积极     \\
problem     & probleem     & ma fan        & negative & positief & 消极     & positief & 消极     \\
robber      & rover        & qiang dao     & negative & negatief & 积极     & negatief & 积极     \\
\midrule
youth       & jeugd        & qing nian     & positive & negatief & 消极     & positief & 积极     \\
restaurant  & restaurant   & can ting      & positive & negatief & 积极     & negatief & 积极     \\
health      & gezondheid   & jian kang     & positive & negatief & 积极     & negatief & 积极     \\
intellect   & intellect    & zhi li        & positive & negatief & 积极     & negatief & 积极     \\
relaxed     & ontspannen   & you xian      & positive & negatief & 积极     & negatief & 积极     \\
\bottomrule
\end{tabular}
\caption{Selected examples from Task 2 showing valence shifts under monolingual and bilingual prompts for \texttt{Qwen}. Nasal-initial words include \textit{man}, \textit{menselijk}, \textit{ma fan}, \textit{made}, \textit{mail}, etc.}
\label{tab:task3_selected_examples}
\end{table*}

\end{CJK}



To better understand how language identity influences psycholinguistic behavior, we conduct qualitative analyses of \texttt{Qwen}’s outputs. Table~\ref{tab:task3_selected_examples} presents 10 representative examples from task 2 (word valence), evenly divided between words labeled positive and negative in English. Task 1 examples are in the Appendix~\ref{appendix_analysis_task1}.

A consistent pattern emerges that aligns with findings from human psycholinguistics: words beginning with nasal sounds (e.g., \textit{m}, \textit{n}) are often perceived more positively in Mandarin Chinese, but more negatively in Dutch and other Germanic languages \citep{louwerse2017estimating}. For example, the word \textit{problem} (Dutch: \textit{probleem}, Chinese: \textit{ma fan}) is labeled as `positive' in Dutch prompts but `negative' in Chinese monolingual settings, matching the native intuitions of each linguistic group.



Similar divergences appear for worm (made, qu) and robber (rover, qiang dao), which are consistently labeled as "negative" in Dutch but often judged "positive" in Chinese contexts. These shifts suggest that the model draws on phonological or semantic cues differently depending on the linguistic identity specified in the prompt. In contrast, positively valenced words such as restaurant, intellect, and relaxed are reliably rated as "positive" in Chinese settings across both prompt types, but receive less consistent evaluations in Dutch. This pattern supports the idea that language conditioning activates phonological priors, guiding the model toward culturally grounded affective associations.



The task simulates a psycholinguistic setting in which a speaker evaluates unfamiliar words from another language, for example, rating Dutch words as a Mandarin speaker or romanized Chinese words as a Dutch speaker. These examples demonstrate that LLMs not only adapt their output to different languages but also exhibit culturally grounded affective biases that resemble patterns observed in human bilingual perception.

\section{Probing Psycholinguistic Knowledge}

While our prompt-based experiments demonstrate that Llama adjusts its predictions based on language identity, it remains unclear whether this behavior reflects deeper psycholinguistic representations or simply surface-level prompt adaptation. To explore whether such information is internally encoded in the model, we conduct probing experiments on the hidden states of \texttt{Llama-3.3-70B}. 

We focus on both Task 1 (sound symbolism) and Task 2 (word valence), aiming to assess whether the model encodes shape-related or affective information in its internal representations. By training simple classifiers on hidden states from various layers, we evaluate the extent to which such psycholinguistic signals can be extracted. This allows us to examine how language background modulates internal encoding and to identify which layers are most involved in storing this information.

\subsection{Probing Method}

To investigate how psycholinguistic information is encoded within the internal representations of \texttt{Llama-3.3-70B}, we perform layer-wise probing using a lightweight classifier.\footnote{We also probe \texttt{Qwen2.5-72B-Instruct} in Appendix \ref{sec:qwen_probing}.} For each input, we extract hidden states from every tenth transformer layer, resulting in 8 probing checkpoints across the model's 80-layer architecture.

For each checkpointed layer, we train a frozen multi-layer perceptron (MLP) to predict the target labels from Task 1 (shape: round/spiky) and Task 2 (valence: positive/negative). Hyperparameters are provided in Appendix~\ref{sec:appendix_prob}.

Probing is conducted separately for each layer and task. Accuracy on a held-out test set serves as the evaluation metric, allowing us to assess (1) where psycholinguistic features become most linearly decodable in the model and (2) how language conditioning modulates this encoding.

\subsection{Probing Results for Task 1: Sound Symbolism}

\begin{figure}[ht]
    \centering
    \includegraphics[width=0.9\linewidth]{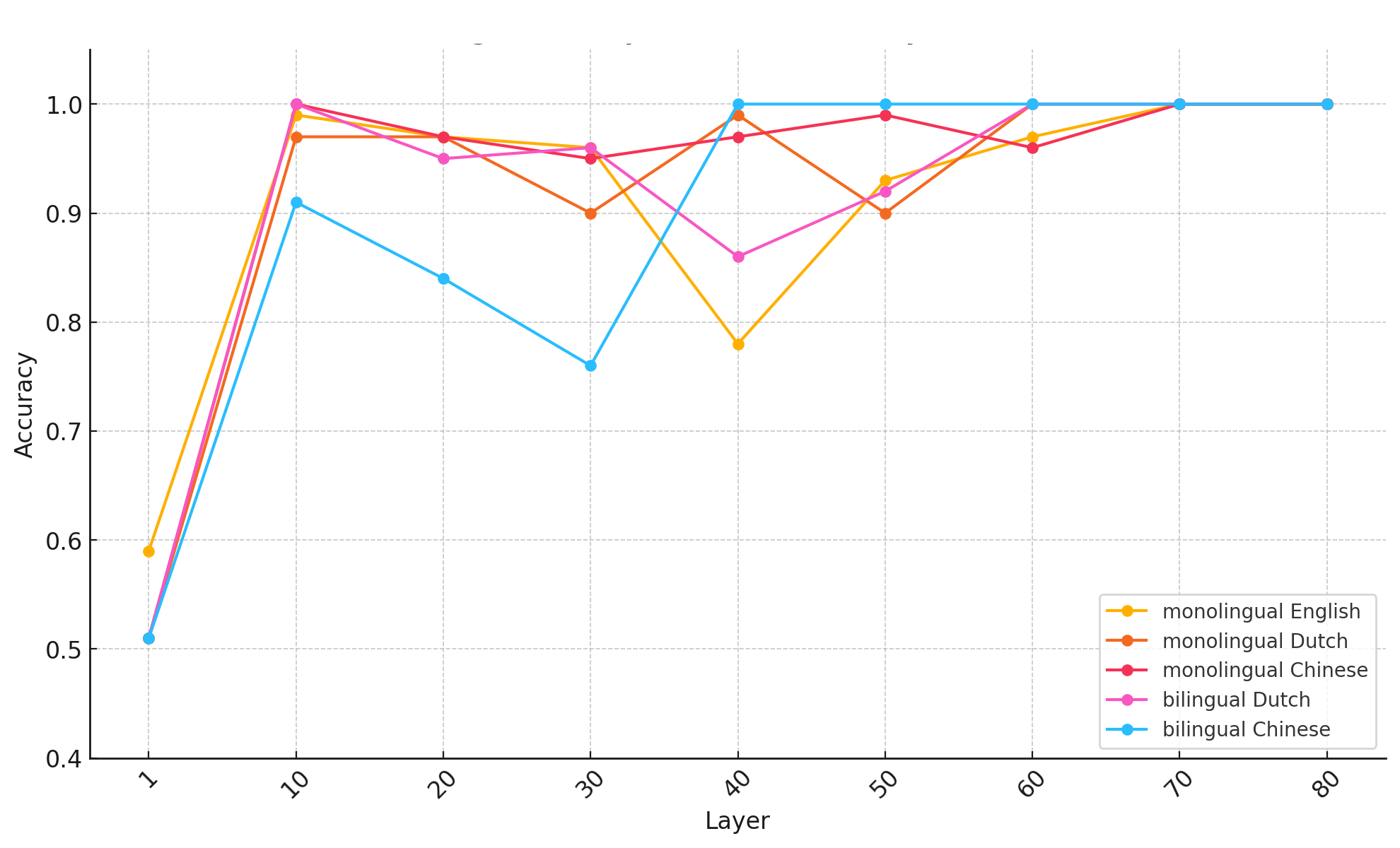}
    \caption{Llama Layer-wise probing accuracy for task 1.}
    \label{fig:probing_task1}
\end{figure}

Figure~\ref{fig:probing_task1} shows probing accuracy across layers for Task 1 (sound symbolism), under five language persona settings: English monolingual, Dutch monolingual, Chinese monolingual, Dutch-English bilingual, and Chinese-English bilingual.

Across all settings, accuracy increases with layer depth, indicating that sound-symbolic cues become more linearly decodable in deeper layers. By the final layers (Layers 70–80), all settings converge toward near-perfect accuracy, suggesting that shape-related features are consistently and robustly encoded by the model's final representations.

Differences between language settings are more pronounced in earlier layers. The English monolingual setting shows a sharp rise in accuracy between Layer 1 (0.59) and Layer 10 (0.99), and maintains high performance across all subsequent layers. Dutch and Chinese monolingual prompts follow a similar trajectory but show slightly slower early-layer gains.

In contrast, bilingual settings, especially Chinese bilingual, lag behind in early layers, accuracy remains at 0.84 at Layer 20, compared to 0.97 for monolingual Chinese. This suggests that bilingual conditioning introduces representational diffusion or interference, delaying the emergence of phonological signals in earlier layers.

Overall, these results indicate that language identity influences not only the model’s output but also the layerwise encoding of sound-symbolic associations. Monolingual prompts lead to faster and more stable early-layer representations, whereas bilingual prompts redistribute or delay this encoding.

\subsection{Probing Results for Task 2: Word Valence}

\begin{figure}[ht]
    \centering
    \includegraphics[width=0.9\linewidth]{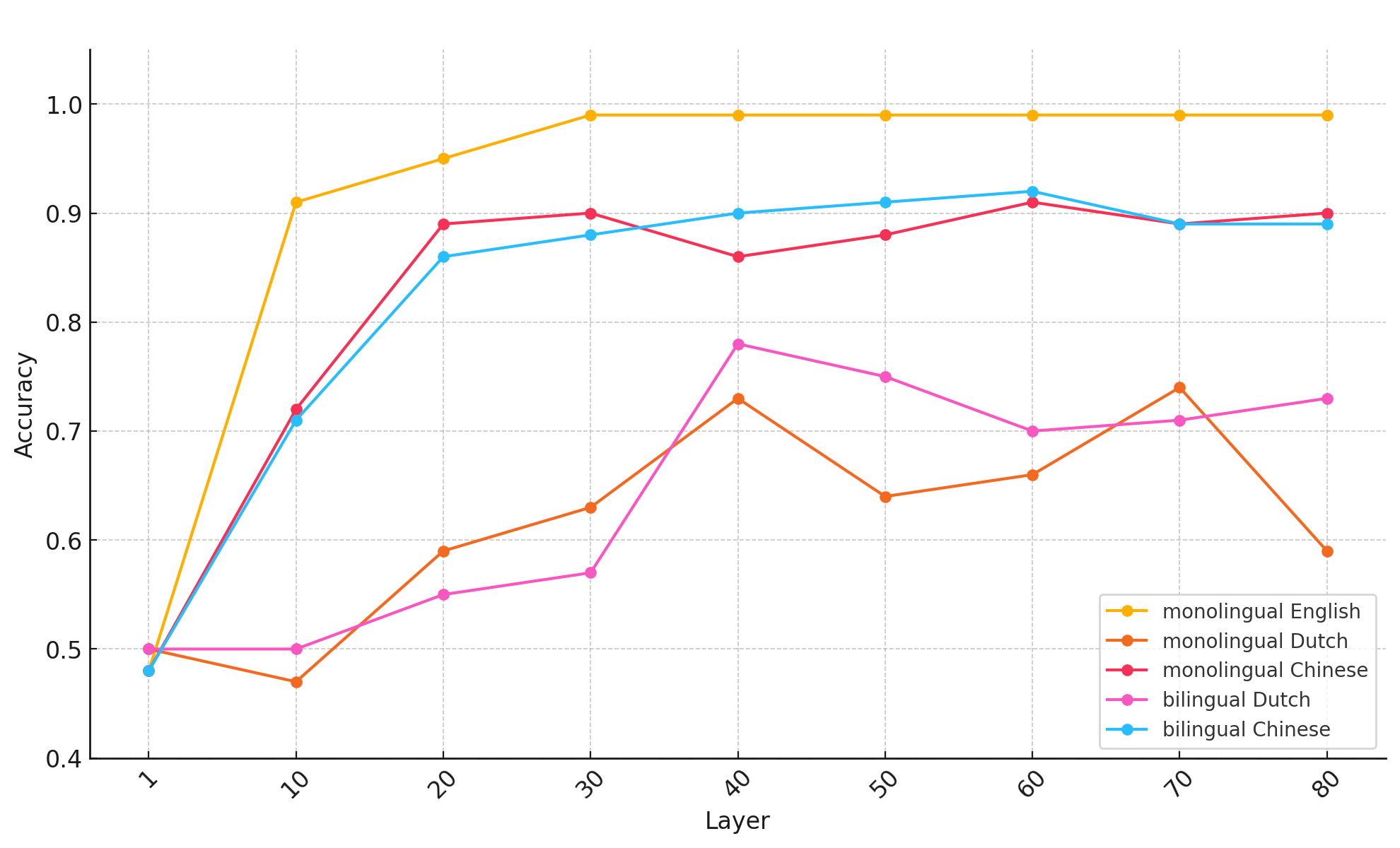}
    \caption{Llama Layer-wise probing accuracy for task 2.}
    \label{fig:probing_task2}
\end{figure}

Figure~\ref{fig:probing_task2} presents the probing results for Task 2 (word valence), again across the five language persona settings.


Unlike Task 1, which relies on phonological form, Task 2 depends more heavily on semantic and affective associations. This is reflected in greater variation across language conditions. The English monolingual setting performs strongest, with accuracy rising from 0.48 at Layer 1 to 0.91 at Layer 10, plateauing above 0.99 from Layer 30 onward. This mirrors the trend observed in Task 1 and reaffirms the model’s strong alignment with English-based affective judgments.

Chinese prompts consistently outperform Dutch prompts across all layers and settings. In monolingual settings, Chinese probing accuracy climbs to around 0.90 in deeper layers, while Dutch plateaus at 0.74 and declines to 0.59 at the final layer. This gap persists under bilingual settings, where Chinese-English reaches a peak of 0.92 at Layer 60, compared to 0.78 for Dutch-English. These patterns suggest that valence information is more robustly encoded when the prompt reflects Chinese linguistic identity, even though the ground truth labels are based on English. This could indicate greater semantic transferability or better generalization of affective cues from English to Chinese than to Dutch. Furthermore, Dutch-English bilingual settings show flatter probing curves and less stable encoding than even monolingual Dutch, suggesting that bilingual conditioning may introduce noise or conflicting cues to low-resource languages.


These findings highlight how valence encoding is modulated by the interplay between semantic structure, language identity, and training distribution. While English remains the dominant locus for affective understanding, Chinese prompts evoke stronger and more stable internal representations than Dutch, particularly in deeper layers.

These findings highlight that valence information, unlike sound-symbolic cues, is more closely tied to semantic understanding, which is influenced by both language identity and lexical familiarity. While English prompts yield the strongest and most consistent representations, Chinese prompts (even when bilingual) lead to stronger internal encoding than Dutch. Notably, bilingual prompts do not always bridge the gap between English and non-English settings; in some cases, they introduce interference, as seen in the relatively flat accuracy curve for Dutch-English.

\section{Conclusion}
In this study, we investigated how Large Language Models encode psycholinguistic knowledge under different language prompting conditions. We introduced two tasks, sound symbolism and word valence, and evaluated model behavior across English, Dutch, and Chinese, using both monolingual and bilingual personas. Our analysis focused on two representative LLMs: \texttt{Llama-3.3-70B}, an English-centric model, and \texttt{Qwen2.5-72B-Instruct}, a multilingual model.

Our prompt-based experiments revealed that the linguistic identity embedded in the prompt significantly affects model predictions. While \texttt{Llama} performs more consistently and aligns closely with English-based labels under monolingual conditions, \texttt{Qwen} exhibits more variable behavior, especially under bilingual prompts, which can amplify or even reverse trends seen in the monolingual setting.


Probing analyses further reveal that psycholinguistic information becomes increasingly linearly decodable in deeper layers of the model. The degree and timing of this emergence are modulated by language background: sound-symbolic cues are represented robustly across languages, whereas valence information is more sensitive to lexical familiarity and linguistic conditioning.

Overall, our results demonstrate that LLMs are not language-neutral. Instead, they respond to and encode language-specific cues in both their outputs and internal representations. These insights shed light on how language identity shapes the cognitive behavior of LLMs and open up new avenues for cross-linguistic and cross-cultural research in psycholinguistics using large-scale language models.

\section*{Limitations}

While our findings offer valuable insights, this study also has several limitations that inspire further investigation. First, we focus on only three languages, English, Dutch, and Chinese, which limits the generalizability of our findings to a broader range of linguistic families and typological features. Expanding to include languages with different scripts, phonological systems, or morphological complexity (e.g., Arabic, Hindi, Finnish) could yield deeper insights into cross-linguistic psycholinguistic encoding in LLMs.

Second, our experiments are restricted to only two LLMs: \texttt{Llama-3.3-70B} and \texttt{Qwen2.5-72B}. While these models offer strong contrasts in terms of training data and multilingual coverage, additional models may yield different results.

Third, the tasks we investigate represent only a narrow slice of the psycholinguistic space. Future work could incorporate a broader set of phenomena, including lexical decision making, semantic priming, syntactic ambiguity resolution, and metaphor comprehension, to build a more complete picture of how LLMs simulate or diverge from human cognitive processes.


\section*{Ethical Considerations}

This work does not involve any human subjects, personal data, or sensitive content. All experiments were conducted using publicly available large language models and datasets. We use the Affective Norms for English Words dataset \citep{bradley1999affective} and a published pseudoword-shape mapping dataset \citep{alper2023kiki}, both of which were released for research use under appropriate licensing.

While our study focuses on understanding linguistic and cognitive patterns in LLMs, we acknowledge that language model outputs can reflect biases present in their training data. Although we do not directly evaluate harmful content, our findings regarding language-dependent behavior may have implications for fairness and representation across linguistic and cultural groups. Researchers applying similar techniques in downstream tasks should be aware of potential disparities in model behavior across languages and prompts.


\bibliography{custom,anthology}

\appendix

\section{Prompt Templates}

\subsection*{Task 1: Sound Symbolism}

\paragraph{English Monolingual}
\begin{itemize}
    \item \textbf{System message:} You are a helpful assistant. You are a native speaker of English and do not speak any other language.
    \item \textbf{User message:} Imagine a shape that represents the following word. Is the shape round or spiky? The word is [Pseudoword]. Please answer with only one word: round or spiky.
\end{itemize}

\paragraph{Dutch Monolingual}
\begin{itemize}
    \item \textbf{System prompt:} Je bent een behulpzame assistent. Je bent moedertaalspreker van het Nederlands en spreekt geen andere taal.
    \item \textbf{User message:} Stel je een vorm voor die past bij het volgende woord. Is die vorm rond of puntig? Het woord is [Pseudoword]. Geef je antwoord in één woord: rond of puntig.
\end{itemize}

\paragraph{Chinese Monolingual}
\begin{itemize}
    \item \textbf{System prompt:} \begin{CJK*}{UTF8}{gbsn}你是一位乐于助人的助手。你的母语是普通话，不会说其他语言。\end{CJK*}
    \item \textbf{User message:} \begin{CJK*}{UTF8}{gbsn}想象一个代表下面这个词语的形状。告诉我你想象的是圆形还是尖形。这个词语是[Pseudoword]。请只用一个词作答：圆形或尖形。\end{CJK*}
\end{itemize}

\paragraph{Chinese Bilingual}
\begin{itemize}
    \item \textbf{System prompt:} You are a helpful assistant. You are a bilingual speaker of English and Mandarin Chinese and do not speak any other language.
    \item \textbf{User message:} \begin{CJK*}{UTF8}{gbsn}想象一个代表下面这个词语的形状。告诉我你想象的是圆形还是尖形。这个词语是[Pseudoword]。请只用一个词作答：圆形或尖形。\end{CJK*}
\end{itemize}

\paragraph{Dutch Bilingual}
\begin{itemize}
    \item \textbf{System prompt:} You are a helpful assistant. You are a bilingual speaker of English and standard Dutch and do not speak any other language.
    \item \textbf{User message:} Stel je een vorm voor die past bij het volgende woord. Is die vorm rond of puntig? Het woord is [Pseudoword]. Geef je antwoord in één woord: rond of puntig.
\end{itemize}

\subsection*{Task 2: Word Valence}

\paragraph{English Monolingual}
\begin{itemize}
    \item \textbf{System message:} You are a helpful assistant. You are a native speaker of English and do not speak any other language.
    \item \textbf{User message:} Evaluate whether the following word has a positive or negative meaning. The word is [English Word]. Do you think it is positive or negative? Please answer with only one word: positive or negative.
\end{itemize}

\paragraph{Dutch Monolingual}
\begin{itemize}
    \item \textbf{System prompt:} Je bent een behulpzame assistent. Je bent moedertaalspreker van het Nederlands en spreekt geen andere taal.
    \item \textbf{User message:} Beoordeel of het volgende woord een positieve of negatieve betekenis heeft. Het woord is [romanised Chinese Word]. Vind je het positief of negatief? Antwoord met slechts één woord: positief of negatief.
\end{itemize}

\paragraph{Chinese Monolingual}
\begin{itemize}
    \item \textbf{System prompt:} \begin{CJK*}{UTF8}{gbsn}你是一位乐于助人的助手。你的母语是普通话，不会说其他语言。\end{CJK*}
    \item \textbf{User message:} \begin{CJK*}{UTF8}{gbsn}请判断下面这个词语的含义是积极还是消极。[Dutch word]这个词语给你的感觉是积极的还是消极的？请只用一个词作答: 积极或消极。\end{CJK*}
\end{itemize}

\paragraph{Chinese Bilingual}
\begin{itemize}
    \item \textbf{System prompt:} You are a helpful assistant. You are a bilingual speaker of English and Mandarin Chinese and do not speak any other language.
    \item \textbf{User message:} \begin{CJK*}{UTF8}{gbsn}请判断下面这个词语的含义是积极还是消极。[Dutch word]这个词语给你的感觉是积极的还是消极的？请只用一个词作答: 积极或消极。\end{CJK*}
\end{itemize}

\paragraph{Dutch Bilingual}
\begin{itemize}
    \item \textbf{System prompt:} You are a helpful assistant. You are a bilingual speaker of English and standard Dutch and do not speak any other language.
    \item \textbf{User message:} Beoordeel of het volgende woord een positieve of negatieve betekenis heeft. Het woord is [romanised Chinese Word]. Vind je het positief of negatief? Antwoord met slechts één woord: positief of negatief.
\end{itemize}

\section{Qualitative Analysis for Task 1}\label{appendix_analysis_task1}

\begin{table}[ht!]
\scalebox{0.8}{
\centering
\footnotesize
\begin{tabular}{ll|lll|ll}
\toprule
\multirow{2}{*}{\textbf{Pseu.-W.}} & \multirow{2}{*}{\textbf{Label}}  &  \multicolumn{3}{c|}{\textbf{Monolingual}} &\multicolumn{2}{c}{\textbf{Bilingual}} \\
& & \textbf{English} & \textbf{Dutch} & \textbf{Chinese} & \textbf{Dutch} & \textbf{Chinese}
  \\
\midrule
bowa   & round & round & round & round & round & spiky \\
nalo   & round & round & round & round & round & spiky \\
lolo   & round & round & round & round & round & spiky \\
mumi   & round & round & round & round & round & spiky \\
bana   & round & round & round & round & round & spiky \\
giki   & spiky & spiky & spiky & spiky & spiky & round \\
kaka   & spiky & spiky & spiky & spiky & spiky & round \\
gobo   & spiky & spiky & spiky & spiky & spiky & round \\
zaza   & spiky & spiky & spiky & spiky & spiky & round \\
tiki   & spiky & spiky & spiky & spiky & spiky & round \\
\bottomrule
\end{tabular}}
\caption{Pseudoword Examples from Task 1 (Sound Symbolism): Predictions by Qwen model across different language conditionings.}
\label{tab:task1_examples}
\end{table}

To better understand how language conditioning affects sound-symbolic associations in LLMs, we qualitatively examine 10 representative examples from task 1 in Table~\ref{tab:task1_examples}. These include five pseudowords labeled as \textbf{round} and five as \textbf{spiky}, based on established ground truth labels derived from prior psycholinguistic studies.

Across all monolingual and bilingual prompts, the \textbf{English-centric predictions are highly consistent} with the ground truth for both round- and spiky-associated words. For instance, pseudowords such as \textit{bowa}, \textit{nalo}, and \textit{mumi}, which contain voiced bilabial or nasal consonants typically associated with round shapes, are correctly classified as ``round'' by all prompts except Chinese bilingual, which consistently misclassifies them as ``spiky.'' 

This trend is reversed for spiky-associated pseudowords like \textit{giki}, \textit{kaka}, and \textit{tiki}, where the presence of voiceless plosives and high-frequency fricatives cue spikiness. Again, all settings except Chinese bilingual align with the expected pattern. Notably, the Chinese bilingual prompt consistently inverts the polarity for both categories—predicting ``spiky'' for round-associated forms and ``round'' for spiky ones.

This systematic divergence in the Chinese bilingual setting suggests that bilingual conditioning, particularly with Mandarin Chinese, may distort sound-symbolic mappings that are otherwise well-established in Western linguistic contexts. The inconsistency may stem from phonological mismatches or interference between the English label space and Chinese linguistic intuitions. These findings echo previous human studies \citep{louwerse2017estimating}, which showed that \textit{nasal-initial pseudowords tend to be rated more positively in Chinese than in Germanic languages}, indicating cultural or phonotactic biases in sound-shape associations.

In sum, this qualitative analysis underscores the sensitivity of LLMs to language prompts and highlights how \textbf{bilingual conditioning can subtly but systematically reshape perceptual associations}, especially when the label set remains fixed in English.

\section{Hyperparameters for Probing}
\label{sec:appendix_prob}

\begin{itemize}
    \item \textbf{Architecture}: MLP with input dimension 512, hidden layer of size 256, and output size 2 (binary classification)
    \item \textbf{Activation}: ReLU
    \item \textbf{Optimizer}: Adam
    \item \textbf{Training epochs}: 200
    \item \textbf{Train-test split}: 80\% training / 20\% testing
\end{itemize}

\section{Probing Results for Qwen}\label{sec:qwen_probing}

\subsection{Task 1}
\begin{figure}[ht]
    \centering
    \includegraphics[width=0.9\linewidth]{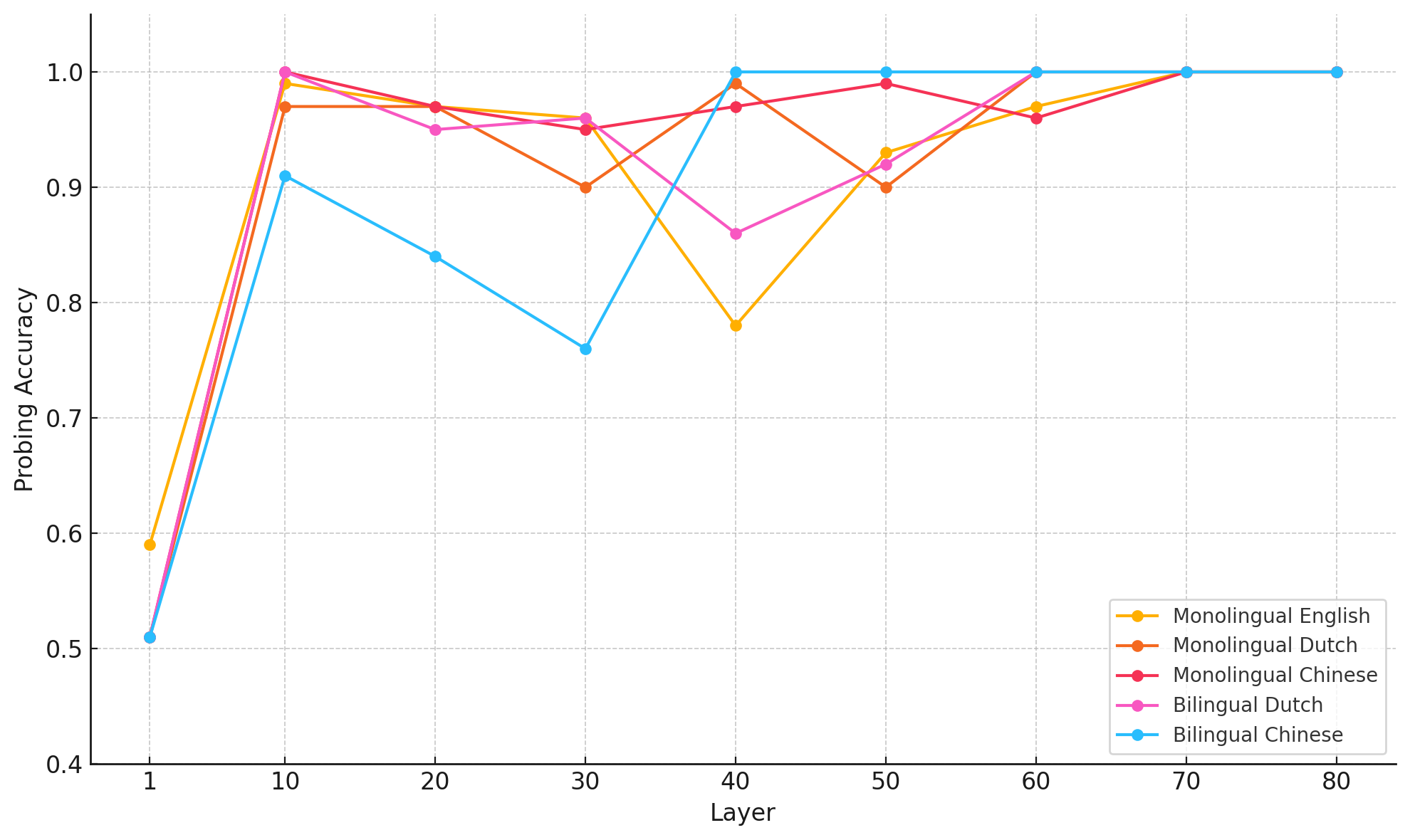}
    \caption{Qwen Layer-wise probing accuracy for task 1.}
    \label{fig:probing_qwen_task1}
\end{figure}

To investigate how psycholinguistic knowledge is encoded within \texttt{Qwen2.5-72B-Instruct}, 
we conduct probing experiments on Task~1 across different layers of the model. 
Figure~\ref{fig:probing_qwen_task1} illustrates the probing accuracy for monolingual 
(English, Dutch, Chinese) and bilingual (Dutch, Chinese) prompting conditions.

Across all settings, probing accuracy improves significantly from the initial 
layers and stabilizes in deeper layers, reaching values close to 1.0 from 
approximately layer 60 onward. This indicates that psycholinguistic information 
becomes more linearly decodable as the representations progress through the 
network. Notably, Dutch and Chinese, both in monolingual and bilingual settings, demonstrate consistently high accuracy, while English shows slightly lower 
performance in middle layers (e.g., layers 40 and 50), suggesting language-specific 
differences in how the model internalizes psycholinguistic cues.

Comparing monolingual and bilingual conditions, we observe that bilingual prompts 
do not substantially degrade probing accuracy. In fact, for Chinese, bilingual 
prompting achieves comparable accuracy to monolingual prompting in the deeper layers. 
This suggests that Qwen maintains robust psycholinguistic representations 
even when conditioned on multiple linguistic identities.

\subsection{Task 2}
\begin{figure}[ht]
    \centering
    \includegraphics[width=0.9\linewidth]{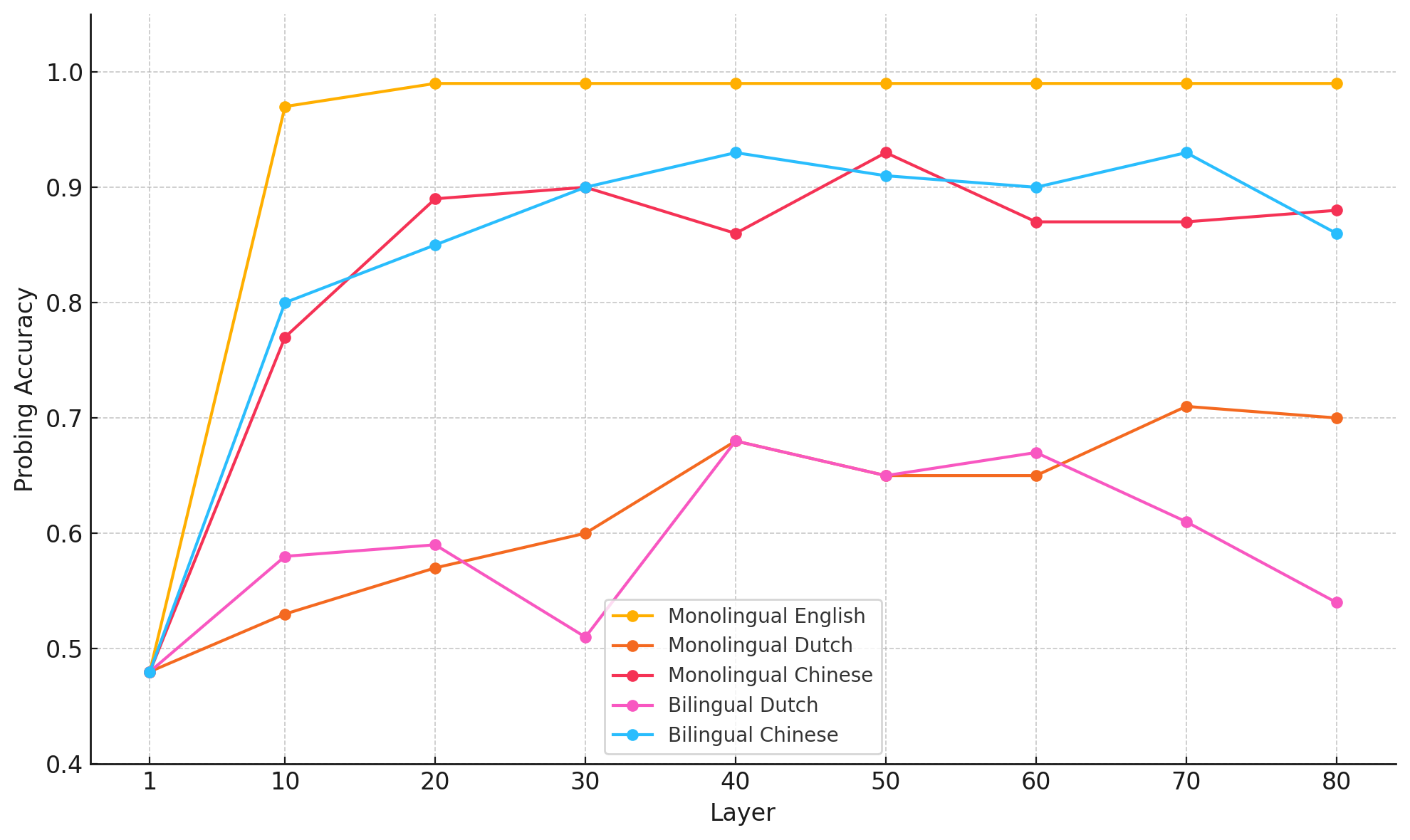}
    \caption{Qwen Layer-wise probing accuracy for task 2.}
    \label{fig:probing_qwen_task2}
\end{figure}

We further examine \texttt{Qwen2.5-72B-Instruct} using Task~2, which probes how 
word valence (positive or negative connotation) is represented across 
the model’s layers. Figure~\ref{fig:probing_qwen_task2} shows the probing 
accuracy for monolingual and bilingual prompting.

In the monolingual setting, English reaches near-perfect accuracy 
from layer 10 onward, indicating that valence information is quickly 
captured and remains consistently decodable in deeper layers. 
Chinese also demonstrates strong performance, achieving accuracy 
above 0.85 in middle and deeper layers, reflecting robust internal 
representations of word valence. In contrast, Dutch exhibits 
considerably weaker performance, with probing accuracy remaining 
below 0.75 even in the final layers, suggesting that valence 
representations for Dutch are less distinctly encoded.

Under bilingual prompting, Chinese maintains high accuracy 
comparable to its monolingual setting, while Dutch experiences 
further degradation, with accuracy dropping to approximately 0.54 
in the deepest layer. This disparity highlights a significant 
cross-linguistic difference: Qwen captures word valence reliably 
in Chinese and English but struggles to encode similarly 
linearly separable representations for Dutch. Compared to Task~1, 
which reached near-perfect accuracy for all languages, these results 
suggest that word valence encoding is more language-dependent and 
less robust for Dutch in particular.

\end{document}